**TITLE:**

# Radiomic Feature Selection Using Gradient Loss of Deep Neural Network for Lung Cancer Stage Detection

**AUTHORS AND AFFILIATIONS:**
Hina Shakir[1], Mohammad Mohatram[2], Javeed Hussain[2], Syed Rizwan Ali[3], Muhammad Irfan Memon[2],
[1] Department of Software Engineering, Bahria University, Karachi, Pakistan;
hinashakir.bukc@bahria.edu.pk
[2] Global College of Engineering and Technology, Muscat, Oman;
m.memon@gcet.edu.om, s.javeedhussain@gcet.edu.om,
[3] Software Engineering & Business Incubation Center, Bahria University Karachi, Pakistan;
rizwanali.bukc@bahria.edu.pk
**Correspondence to**: m.mohatram@gcet.edu.om, s.javeedhussain@gcet.edu.om,
m.memon@gcet.edu.om

**SUMMARY:**
Presented here is a deep learning–based feature selection method that leverages gradients of a neural network loss function with respect to input features to identify and prioritize those that most strongly influence lung cancer stage detection.

**ABSTRACT:**
Radiomics enables extraction of quantitative imaging biomarkers from medical images and has become an important tool for computer-aided cancer diagnosis. However, radiomics datasets are typically high-dimensional with limited samples, making feature selection a critical step for building reliable predictive models. This study proposes a Gradient-Loss Recursive Feature Elimination (GL-RFE) framework that integrates gradient sensitivity analysis from a deep neural network to identify the most influential radiomic features for lung cancer stage detection. A total of 106 radiomic features were extracted from chest Computed Tomography (CT) scans using the PyRadiomics extension of the 3D Slicer platform. The proposed method evaluates feature importance by computing gradients of the network loss with respect to input features and recursively eliminates features with minimal contribution. The resulting top-15 radiomic features are used to train a deep neural network classifier for distinguishing early-stage and advanced-stage lung cancer. The proposed framework achieves strong classification performance, with accuracy of 90.22%, precision of 90.10%, recall of 90.24%, and F1-score of 90.16% on the test dataset. Visualization analyses, including correlation heat maps and distribution plots, further confirm reduced feature redundancy and improved class separability. Compared to conventional feature selection techniques, GL-RFE effectively captures nonlinear feature interactions and enhances model generalization. The presented protocol provides a reproducible and interpretable methodology for radiomics-based cancer stage detection and is particularly suitable for high-dimensional, small-sample biomedical datasets, with potential applications in other domains such as genomics and multimodal clinical analysis.

**INTRODUCTION:**

Lung cancer remains one the major types of cancer leading to serious health concern often leading to death[1]. Radiomics enables quantitative characterization of medical images by extracting large sets of features that describe tumor shape, texture, and intensity patterns[2,3]. These features also termed handcrafted features serve as potential biomarkers for diagnosis, prognosis, and treatment response of lung cancer. However, radiomics datasets are typically high-dimensional and sample-limited, leading to redundant and noisy features that degrade model performance[5,6,7]. Hence, efficient and explainable feature selection is crucial for developing robust radiomics-based predictive models.

Traditional feature selection approaches such as filter methods (e.g., correlation analysis, ANOVA(Analysis of Variance), mutual information) and wrapper methods (e.g., Sequential Feature Selection, Recursive Feature Elimination) are widely used for radiomics-based cancer detection models [4,8,9]. However, they often fail to capture nonlinear feature interactions and deep contextual dependencies inherent in radiomics data[9,10,11]. Ensemble feature learning techniques have been explored for medical image classification but have achieved moderate accuracy, which could be further improved[12].

Deep learning methods, particularly deep neural networks (DNN) have demonstrated superior ability to model nonlinear and hierarchical relationships between features and outcomes, making them ideal for guiding feature selection, thus providing accurate cancer detection models[13,14]. In this context, potential of using Convolutional Neural Network, multimodal AI techniques and VCG-16, a pre-trained models for cancer diagnosis is explored [1,15,16]. A hybrid deep learning model [17] including pre-trained model VGG-19 and long short-term memory networks (LSTMs) was proposed which was trained and tested on large number of images, achieved an accuracy of over 99% accuracy.

Besides cancer detection, there have been research studies conducted for the detection of cancer stage. The authors[18] designed a feed-forward neural network model on NSCLC data base of 300 patients to classify cancer stage I, II and III with an accuracy of 74.52% in model testing. A radiomic-based Bayesian Inversion method[3] is presented for lung cancer stage detection using NLST dataset on a sample size of 200.The proposed method achieved an accuracy of 86%. The literature review has revealed that most of the studies conducted for cancer detection have only performed classification of benign and malignant tumors and only a few studies addressed the classification of cancer stage with an accuracy less than 90% which can further be improved. This research paper addresses the aforementioned research gap and has proposed a robust radiomic features based classification framework for accurate lung cancer stage detection.

A Gradient Loss based Recursive Feature Elimination (GL-RFE) framework has been introduced in this study, which integrates gradient backpropagation from neural networks into the RFE process.

Unlike conventional RFE methods that rely on static feature importance metrics, GL-RFE leverages the gradients of the loss function with respect to each input feature to measure how strongly each feature influences model predictions. By iteratively removing features with minimal gradient contributions, the presented model performs feature selection of top 15 diagnostic features optimized for lung cancer stage detection of 2 classes (stage I & stage II combined and stage IIIa & IIIb combined). The workflow diagram of the work carried out is given in Figure 1. The chosen lung cancer dataset for the presented model is NSCLC Radiomics [19] having 411 volumes of format termed as Digital Imaging and Communications in Medicine (DICOM) with clinical cancer stage information. A total of 106 3D radiomics features of each lung cancer DICOM volume are extracted using PyRadiomics[20], an extension of an open source software 3D Slicer[21]. These features belong to seven features classes[22] including Shape, Gray level Difference Method(GLDM), Gray-Level Co-Occurrence Matrix (GLCM), First Order, Gray Level Run Length Matrix (GLRLM), Gray Level Size Zone Matrix (GLSZM), Neighborhood Gray-Tone Difference Matrix NGTDM. The radiomics data of minority class (stage I & stage II) was oversampled using Synthetic Minority Over-Sampling Technique (SMOTE)[23].

The novelty of the proposed GL-RFE framework lies in integrating gradient-based sensitivity analysis derived from neural network training into the recursive feature elimination process. Unlike conventional RFE approaches that rely on static importance measures, GL-RFE dynamically evaluates feature relevance through back-propagated gradients of the loss function. This enables identification of features that directly influence model predictions for cancer stage while maintaining interpretability and computational feasibility for relatively small medical datasets.

The aforementioned deep learning model with GL–RFE method is implemented in Jupyter notebook in Google Colab. Using the presented method in the Protocol section, the top 15 radiomic features for lung cancer detection are identified and are used for accurate cancer detection on test datasets.

**PROTOCOL:**

The dataset employed for the training and testing of proposed framework is a publicly available CT images dataset of 422 lung cancer patients known as NSCLC Radiomics[17]. For each patient, the data set includes a CT volume and a DICOM Radiotherapy Structure Sets (RTSTRUCT) and DICOM Segmentation (SEG) file. These files contain manual delineations performed by a radiation oncologist of the three-dimensional volume of the primary gross tumor volume (GTV-1), as well as the lung image. The data set is a pre-processed set and the dimensions of the images are 512 x512 pixels.

Due to nonlinear nature of handcrafted radiomic features, the extracted features cannot be directly used with deep learning models for cancer diagnosis and their inherent data patterns need to be captured using AI techniques. GL-RFE ranks features based on the magnitude of the gradient of the model loss L with respect to each input feature $x_i$.

For each input feature $x_i$, the average absolute gradient is calculated as follows:

$$I(x_i) = \frac{1}{N}\sum_{j=1}^{N}\frac{\partial L_j}{\partial x_{ij}} \qquad (1)$$

Here N is the total number of samples. $L_j$ is the loss for $j^{th}$ sample. $x_{ij}$ is the $i^{th}$ feature of the $j^{th}$ sample. $\frac{\partial L_j}{\partial x_{ij}}$ represents sensitivity of the loss to the input feature.

The features with low gradient magnitudes have minimal impact on model updates and are recursively eliminated. The proposed workflow for eliminating the low gradient radiomic features using a Multi-Layer Perceptron (MLP) and training a Deep Learning Neural Network (DNN) on the top 15 features is shown in Figure 1. The performance of GL-RFE method for feature selection is then evaluated.

The proposed model has been written in Jupyter Notebook accessible through Google colab which allows to write Python code and execute in online environment. Different packages which are mentioned the Protocol steps as well as Material need to be downloaded to compile the code.

## 1. EXTRACTION OF RADIOMIC FEATURES USING 3D SLICER PyRadiomics EXTENSION

The following steps are designed to compute radiomic features of a lung CT DICOM file using 3D Slicer PyRadiomics extension and to save in a file of comma separated value (csv) format.

1.1. **Install 3D Slicer**
Install and open 3D Slicer (use the latest stable release from https://download.slicer.org/.

1.2. **Install the PyRadiomics Extension and RT Slicer**
In the menu bar, go to View → Extensions Manager. Then, search for "Radiomics" or "SlicerRadiomics" and RT Slicer. Click Install and it will install RT Slicer and PyRadiomics libraries. Restart 3D Slicer after installation.

1.3. **Download NSCLC RADIOMICS**
Download the CT lung DICOM datasets along with SEG, RTSTRUCT files of 422 patients from https://www.cancerimagingarchive.net/collection/nsclc-radiomics/

1.4. **Load Lung CT DICOM Data**
Go to DICOM module. Click Import and select the folder containing DICOM CT slices and its SEG file in RTSTRUCT modality. After import, double-click the patient/study/series to load it into the Slicer scene. The 3D CT volume should be visible in the viewer panel as shown in Figure 2.

1.5. **Check Geometry Alignment**
In the Data module, expand both CT volume and segmentation. The segmentation should sit exactly over the CT (no misalignment).

1.6. **Open the "Radiomics" Module**

Select Radiomics module from the module section (or search it in the module search bar). In the Input Image Volume, select the desired CT volume. In Input Label/Segmentation, select the segmentation node (the ROI).

1.7. **Adjust Extraction Customization Parameters**
Set Resampled pixel spacing = [1,1,1] (ensures isotropic voxels) and Bin width = 25 (standard for CT). Set LoG Kernel size =2.0, 3.0, 4.0, 5.0

1.8. **Run Feature Extraction**
Click Apply. Software will now compute 3D first-order, shape, and texture features (GLCM, GLRLM, GLSZM, GLDM, and NGTDM). Display the table to verify the computed tables as Figure 3. Output a csv file with all extracted features. Do the above process for all DICOM volumes downloaded from NSCLC RADIOMICS dataset and save it as single file "radiomics.csv".

## 2. DEVELOPING A RADIOMICS-BASED CANCER DETECTION MODEL USING PYTHON LIBRARIES

Following steps are summarized for a user to develop, train and test a cancer detection model using Python Libraries with radiomic features of CT datasets.

2.1. Format the radiomics dataset saved in csv format such that each row should represent one patient/sample, and each column should represent a feature. Include one label column for class labels.
2.2. Open a new Jupyter notebook in Colab environment and start writing code by declaring given below function definitions and built-in Python functions in step 2.3.
2.3. Give command to install Pytorch, torchvision, scikit-learn, numpy, pandas matplotlib, imbalanced-learn on top of the Jupyter notebook.
2.4. Write a function files.upload() to take input csv file from user and store in X,y variables.
2.5. Normalize the stored data using function scaler = StandardScaler() and scaler.fit_transform().
2.6. Define a Multi-Layer Perceptron function MLP(nn.Module()) with configurable hidden layers
2.7. Define a training function def train_epoch() to compute back propagation loss from MLP model with inputs.
2.8. For a number of epochs, define a function def compute_input_gradients() to compute mean gradients of the loss w.r.t. input features and iteratively drop the feature with low gradient i.e. having less importance till 15 features are left.
2.9. Split the data in the ratio of 80% 20% using train_test_split() function with selected 15 features. Apply a 5-fold cross validation using StratifiedKFold(n_splits=5) on the training data to ensure robustness.
2.10. Create large MLP neural network final_model = DNN().
2.11. Evaluate the performance of trained model using test data with following functions: def plot_confusion_matrix(), accuracy_score(), precision_score(), recall_score(), f1_score(), heatmap().

## 3. RUNNING THE JUPYTER NOTEBOOK TO BUILD AND TEST MODEL

3.1. Run the Python code in Jupyter notebook. A prompt is received to upload the radiomics csv file as shown in Figure 4.
3.2. Upload the radiomics.csv
3.3. Save the classification results and generated graphs.

## REPRESENTATIVE RESULTS:

### DATASET SUMMERY

The NSCLC Radiomics dataset comprises 422 CT volumes of lung cancer stage I, II, III patients. While the number of CT datasets with early stage cancer (I, II) is 134, the data samples with advanced stage cancer (IIIa, IIIb) are 288. The dataset exhibited a significant class imbalance, with a higher number of advanced-stage (stage III) cases compared to early-stage (stage I & stage II) cases. To address this imbalance, over-sampling was applied on the extracted radiomic features to increase the representation of the minority class. As a result, the number of Stage I & II samples increased substantially, leading to a more balanced distribution across classes demonstrated in Table 1. This adjustment helps ensure a more reliable and unbiased evaluation of the model's performance.

### RADIOMIC FEATURES EXTRACTION

The workflow of the proposed framework is represented in Figure 1. The step 1.4 and step 1.5 in the Protocol section describe loading CT volume and segmentation mask in 3D Slicer and display of their alignment. The result of the above two steps is shown in Figure 2. Then, radiomic features of the CT volumes were extracted using Radiomics module of 3D Slicer, as shown in step 1.8 of Protocol. The result of using 3 steps from 1.6 till 1.8 of Protocol section is shown in Figure 3 where radiomic features are extracted from seven feature classes using Radiomics module. Details of these feature classes and extracted features are given in Supplementary File (File A).

### GRADIENT LOSS BASED RADIOMIC FEATURE ELIMINATION METHOD

The gradient loss based radiomic feature elimination method (GL-RFE) is executed using Python code implemented in step 2 of the Protocol. The 'radiomics.csv' file which contains radiomic features and cancer stage data of cancer patients is uploaded through a prompt when the written Python code is run. This step is demonstrated through Figure 4.

The hyper-parameters setting of MLP and DNN used for feature elimination and training on the top 15 radiomic features respectively are reported in Table 2. To ensure robustness of the proposed framework**,** k-fold cross-validation(CV) (k=5) was performed on the training dataset. In each fold, the model was trained on 80% of the data and validated on the remaining subset. The feature selection process was executed within each fold to avoid data leakage. The mean accuracy, precision, recall, and F1-score across all folds for cancer stage detection of stage I& II and stage IIIa & IIIb were then computed to assess model stability. The achieved accuracy, precision, recall and F1 score of the developed model in the Protocol section for test dataset is 90.22%, 90.10%, 90.24% and 90.16% respectively which is shown along with top 15 diagnostic features in Figure 5. Out of 42 patients with cancer stage I&II radiomics data, 38 were detected correctly with the model. From 50 patients with stage IIIa & IIIb radiomics data, developed model predicted 45 correctly as demonstrated in the confusion matrix in Figure 6. The results showed consistent performance across folds, indicating that the GL-RFE feature selection process generalizes well and is not dependent on a specific data partition.

The proposed framework performed was analyzed through several visualizations. A comparison of correlation heatmap of all the extracted radiomic features and selected radiomic features is shown in Figure 7. Kernel density estimation plots of top 6 features were illustrated to further illustrate feature separability in Figure 8. Furthermore, a radar plot was used to visualize evaluation metric performance in a single figure in Figure 9, where non-overlapping density curves indicate strong discriminative capability and clear class boundaries between cancer stages.

**FIGURE AND TABLE LEGENDS:**

Figure 1: Workflow diagram of proposed method, where radiomic features with low mean gradients of the back propagation loss are iteratively dropped and a DNN is trained on top 15 features for cancer detection

Figure 2: Display of a 3D CT volume along with its segmentation mask from NSCLC Datasets when DICOM file and its segmentation file are loaded in 3D Slicer

Figure 3: Screenshot of computed radiomic features of segmented Lung Tumor using Radiomics module in 3 –D Slicer.

Figure 4: A prompt window to select radiomics.csv file containing computed radiomic features from step 1 shows up when Python code for lung cancer detection in step 2 is run.

Figure 5: Display of Evaluation metric including accuracy, precision, recall and F1-score of GL-RFE method when trained DNN is tested using Python code developed in step 2.

Figure 6: Computation of Confusion Matrix of the test datasets showing diagnosis of early stage and advanced-stage lung cancer

Figure 7: Heatmaps of 106 computed radiomic features and selected 15 features; showing correlation in the computed features

Figure 8: Kernel density estimation (KDE) plots to provide insight into class-wise feature distributions to highlight their strong discriminative capability.

Figure 9: Radar plot analysis of evaluation metrics

Table 1: Cancer Stage wise CT Data Sets Summary before and after over-sampling to address class imbalance problem

Table 2: Hyper-parameters setting used in training of MLP and DNN for data training

Table 3: Computational complexity comparison of common feature selection techniques and the proposed GL-RFE method.

Table 4: Performance Comparison of GL-RFE with classical machine learning techniques

**DISCUSSION:**

The robustness and reliability of the proposed framework are evident from the high values of evaluation metrics including accuracy, recall, precision and F-1 score [24]. All scores obtained over 90% performance on the test data with 5-fold CV employed during MLP training**.**

The performance and validity of the proposed GL-RFE framework were further supported through visualization techniques. Correlation heatmaps [25] in Figure 7 reveal that the initially extracted radiomic features exhibit substantial inter-feature redundancy, whereas the selected feature subset demonstrates significantly reduced correlation, indicating effective elimination of redundant information and improved feature independence. Furthermore, Kernel density estimation (KDE) plots[26] in Figure 8 provide insight into class-wise feature distributions, where several selected features show reduced overlap between early-stage and advanced-stage cancer groups, highlighting their strong discriminative capability. Additionally, radar plot analysis[27] of evaluation metrics was performance in Figure 9 which demonstrates a balanced and consistent performance across accuracy, precision, recall, and F1-score, forming a near-uniform shape that reflects the robustness and generalization ability of the model. Together, these visualization-based analyses confirm that the proposed method not only improves predictive performance but also enhances interpretability and reliability of radiomics-based cancer detection.

A comparison of the computational complexity[28] of commonly used feature selection methods with the proposed GL-RFE framework is drawn in Table 3. Filter-based methods such as ANOVA and mutual information exhibit low computational complexity since features are evaluated independently. Wrapper approaches such as SVM-RFE require repeated model retraining, which significantly increases computational cost as the number of features grows. The proposed GL-RFE method relies on gradient information obtained during neural network training, resulting in complexity proportional to the number of epochs and features. Although computationally more intensive than simple filter methods, GL-RFE provides improved feature relevance estimation by capturing nonlinear dependencies within radiomic data.

Due to limited accessibility of lung cancer datasets with cancer stage information, the study could not be validated on another lung cancer dataset. However, a comparison of proposed method with both classical machine learning and deep learning approaches using all extracted features has been reported in Table 4. Conventional models[5] such as Support Vector Machine (SVM), K Nearest Neighbors(KNN), Logistic Regression show comparatively lower performance due to their inability to effectively handle redundant and high-dimensional radiomic features. Ensemble methods[7] such as Random Forest improve performance by modeling nonlinear relationships; however, they still rely on the complete feature set. Deep learning approaches, including Convolutional Neural Network (CNN)[15] and autoencoder-based model[30], demonstrate improved performance by capturing complex feature interactions and latent representations. The proposed GL-RFE framework demonstrates a consistent improvement over baseline models. When compared with conventional machine learning methods such as SVM and Random Forest, the improvement is more pronounced, reaching up to +4.02% in accuracy over SVM and +3.22% over Random Forest. These gains highlight the effectiveness of gradient-based feature elimination in removing redundant radiomic features and enhancing discriminative learning. The ANN model trained on all features achieves 87.10% accuracy, which is 3.1% lower than GL-RFE, highlighting the negative impact of redundant and irrelevant features on model performance. The deep learning techniques including CNN and autoencoder successfully predicted the lung cancer stage, however, the performance of proposed method is superior to deep learning model.

The comparative results of proposed method with other classic feature selection techniques in Table 5 demonstrate a clear performance progression across different categories of feature selection methods. Filter-based techniques5 such as ANOVA, Chi-square, and Mutual Information achieve accuracies in the range of approximately 84.5%–85.8%, indicating limited discriminative capability due to their inability to capture inter-feature dependencies. Embedded methods[7] show moderate improvement, with LASSO achieving 86.70% accuracy and Random Forest reaching 87.90%, reflecting the benefit of incorporating model-based feature importance. Wrapper methods[7] further enhance performance, where Sequential Feature Selection (SFS) and conventional Recursive Feature Elimination (RFE) achieve 86.75% and 88.30% accuracy, respectively, demonstrating the advantage of iterative feature evaluation. In comparison, the proposed GL-RFE framework achieves the highest performance, with 90.22% accuracy, 90.10% precision, 90.24% recall, and 90.19% F1-score, outperforming the best baseline (RFE) by approximately 1.9% in accuracy and ~1.8% in F1-score. Notably, GL-RFE also exhibits more balanced metric values, with a maximum variation of less than 0.3% across precision, recall, and F1-score, indicating stable and consistent classification performance.

Overall, the quantitative improvements demonstrate that gradient-based feature selection provides a measurable advantage over traditional methods by effectively capturing nonlinear relationships and reducing feature redundancy. The consistent gain across all evaluation metrics, rather than improvement in a single metric, further confirms the robustness and generalization capability of the proposed GL-RFE framework.

The successful execution of the GL-RFE protocol relies on several critical steps that directly influence the reliability and reproducibility of the results. First, accurate segmentation and proper alignment between CT volumes and region-of-interest masks during radiomic feature extraction are essential, as misalignment can lead to erroneous feature computation[2,5]. Second, consistent preprocessing particularly normalization and standardization of radiomic features is crucial to ensure stable neural network training. Third, the design and training of the deep neural network significantly impact gradient estimation; appropriate selection of hyper-parameters such as learning rate, number of hidden layers, and number of epochs is necessary to obtain meaningful gradient signals. Finally, performing feature selection within each fold of cross-validation is a key methodological step to prevent data leakage and ensure unbiased evaluation, particularly in small-sample radiomics datasets [6,8].

Several modifications and troubleshooting strategies can enhance the robustness of the proposed technique. In cases of unstable or noisy gradients, smoothing strategies such as averaging gradients across multiple epochs or introducing regularization (e.g., dropout or weight decay) can improve feature importance estimation. If class imbalance affects model performance, techniques such as SMOTE, as used in this study, or alternative resampling strategies can be employed to stabilize training. Additionally, reducing feature elimination step size or using early stopping criteria can prevent the removal of moderately important features during recursive elimination. For datasets with high heterogeneity, normalization parameters and bin width settings in radiomics extraction may require adjustment to ensure feature consistency across samples[4,9]. These practical considerations make the protocol adaptable to varying dataset characteristics.

Despite its advantages, the GL-RFE framework has a few limitations. The method is computationally intensive due to repeated neural network training and gradient computation, particularly for large-scale datasets or high-dimensional feature spaces. Moreover, gradient-based importance reflects local sensitivity of the loss function and may not fully capture global or causal relationships between features and outcomes, unlike explainability methods such as SHAP or LIME[29]. The approach is also sensitive to hyper-parameter settings, which can influence feature ranking stability.

The significance of the presented technique extends beyond improved predictive performance. By identifying a compact set of highly informative radiomic features, the method enhances model interpretability and reduces overfitting, which are critical requirements for clinical deployment of AI-based diagnostic tools [3,5]. Furthermore, the integration of deep learning with handcrafted radiomic features bridges the gap between traditional radiomics and modern Artificial Intelligence-driven approaches, providing a hybrid framework that is both explainable and effective. This makes GL-RFE particularly relevant for precision medicine applications where understanding feature relevance is as important as prediction accuracy.

## CONCLUSION

This study presented a gradient-loss-based recursive feature elimination framework for identifying diagnostically relevant radiomic features for lung cancer stage detection. By integrating gradient sensitivity analysis from deep neural networks with feature elimination, the proposed approach effectively identifies the most informative imaging biomarkers while reducing redundancy within the radiomic feature space. Experimental evaluation on the NSCLC Radiomics dataset demonstrates strong classification performance and highlights the capability of gradient-based feature selection to capture nonlinear relationships among radiomic features. The proposed framework provides a promising methodology for improving interpretability and generalization of radiomics-based diagnostic models. Future work will focus on extending the proposed GL-RFE framework to multi-class cancer stage classification and integration with deep feature representations extracted directly from medical images. In addition, hybrid feature selection strategies combining gradient sensitivity with explainable AI techniques such as SHAP or LIME may further enhance interpretability of radiomics models. Another promising direction involves incorporating multimodal data including genomic, clinical, and imaging features to improve predictive accuracy and support personalized cancer treatment planning.

**ACKNOWLEDGMENTS:**

Not Applicable

**DISCLOSURES**
The authors declare that they have no competing financial interests.

## Initialize MLP

Train a neural network on all features.

architecture:

Input → Dense (128) → ReLU → Dense (64) → ReLU → Dense (1) → Sigmoid

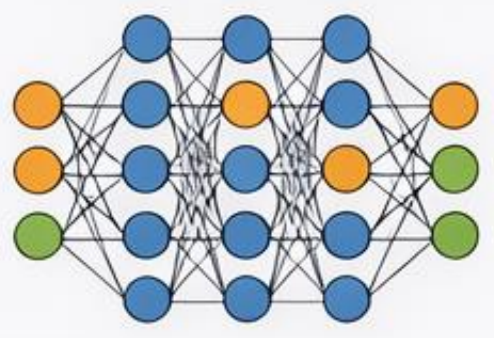

**1.** Train on 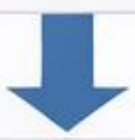 **All Features**

## Compute Gradient Importance

Calculate mean absolute gradient for each feature:

$$I(x_i) = \frac{1}{N} \sum_{j=1}^{N} \left| \frac{\partial L}{\partial x_{i,j}} \right|$$

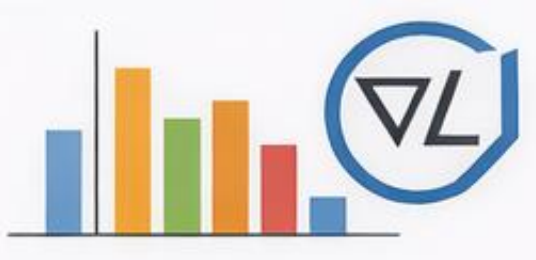

**2.** Compute **Gradients** 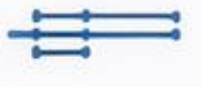

## Rank and Eliminate

Sort by gradient importance and remove features with lowest gradient.

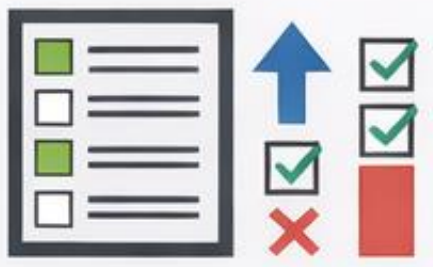

**3.** Remove Least Important Features

## Retrain

Re-train and recompute gradients on remaining features.

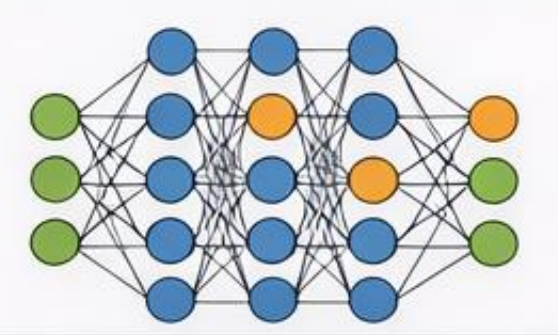

**4.** Re-Train Model

## Iterate

Repeat until 15 features remain.

**Feature Selection Iteration** →

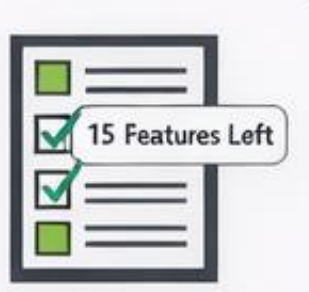

**6.** Train and Evaluate Model

## Train DNN

Train a DNN with the top 15 features and evaluate performance.

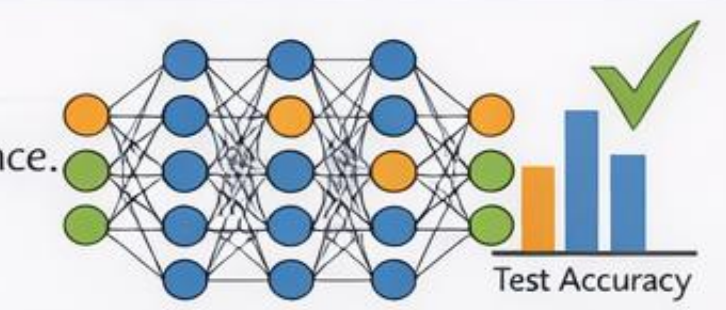

Figure 1

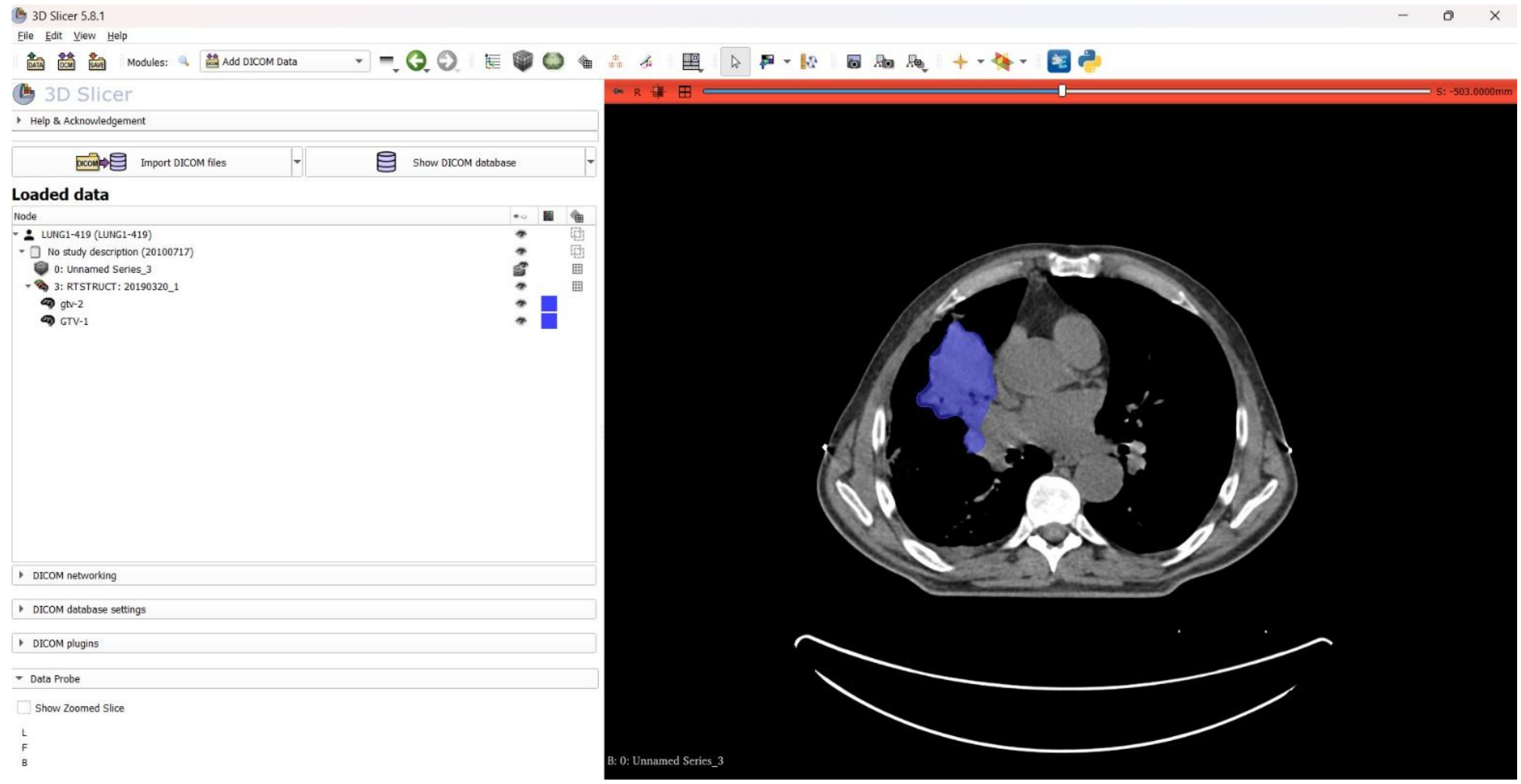

Figure 2

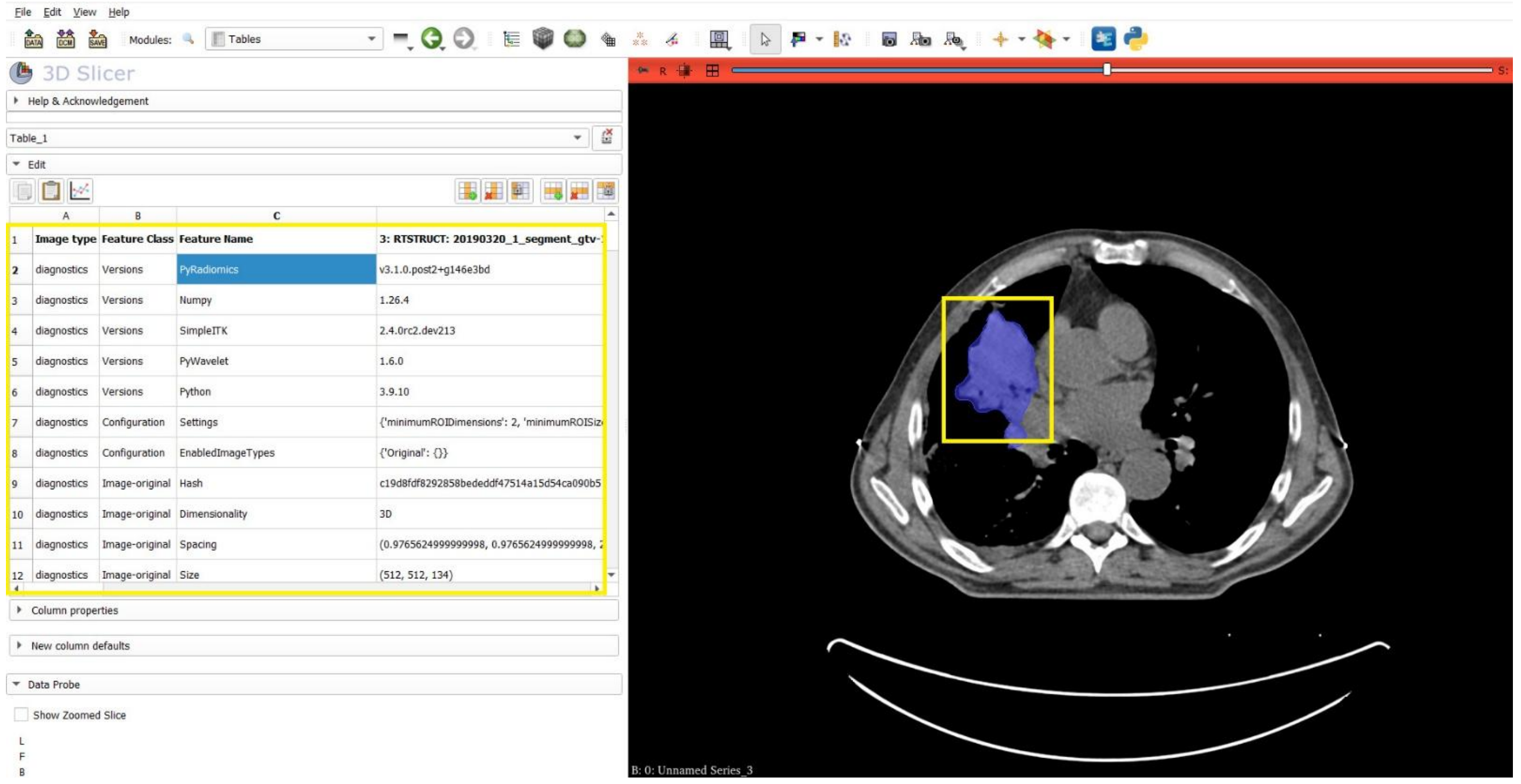

Figure 3

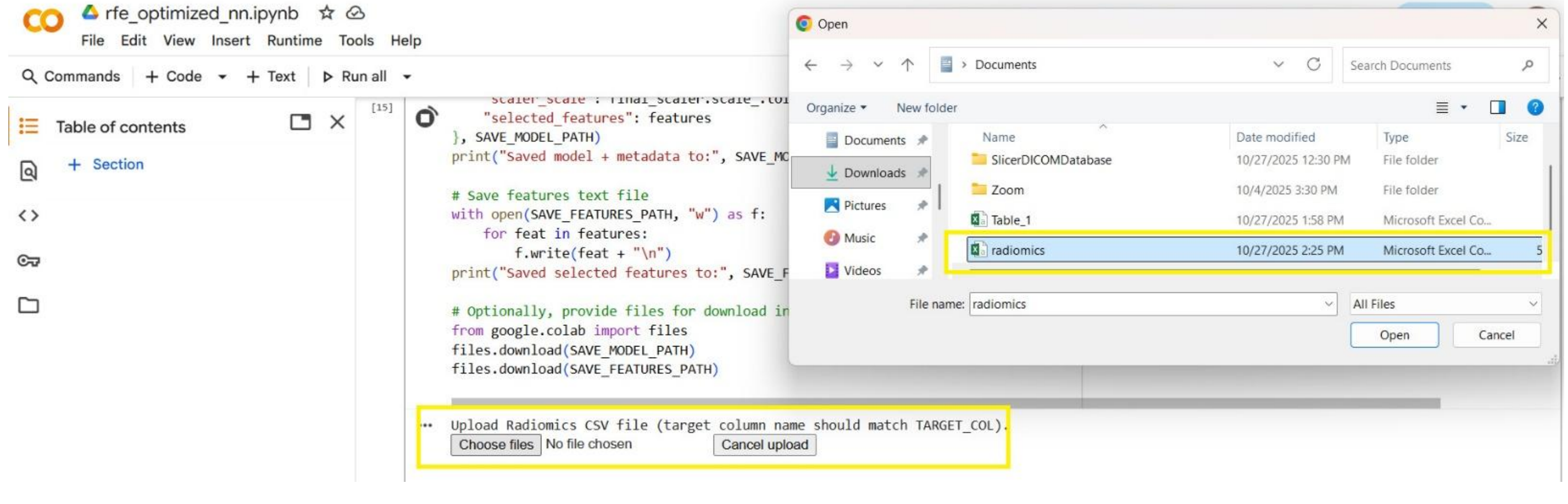

**Figure 4**

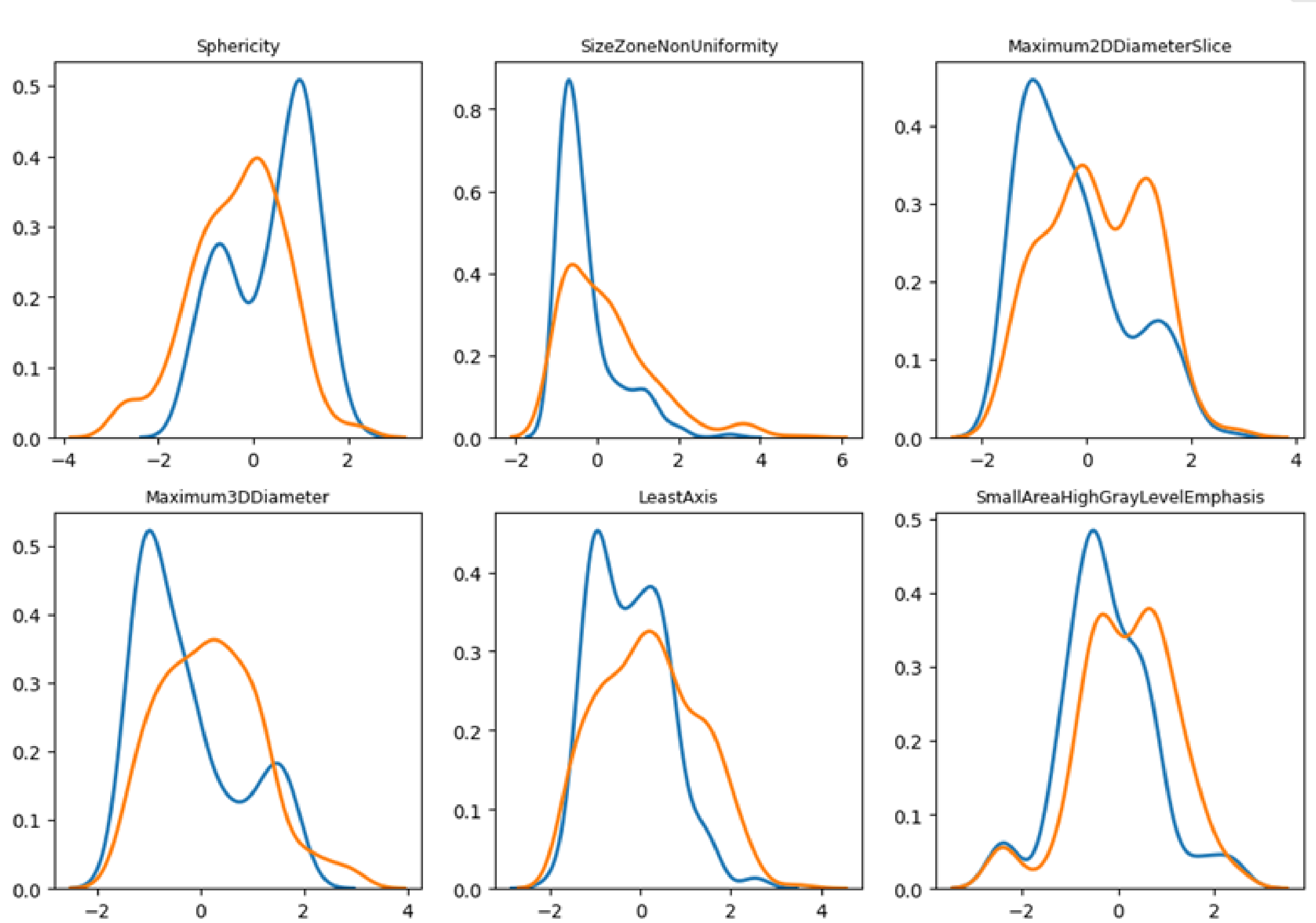

**Figure 5**

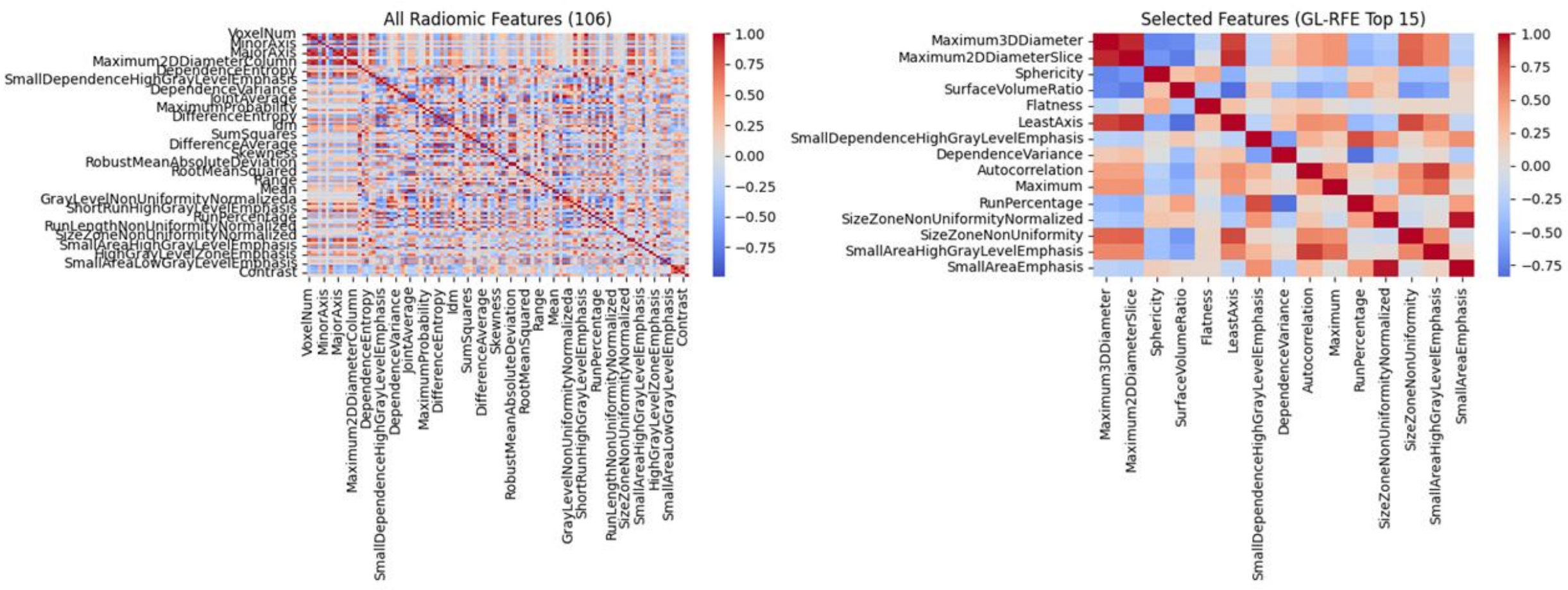

Figure 6

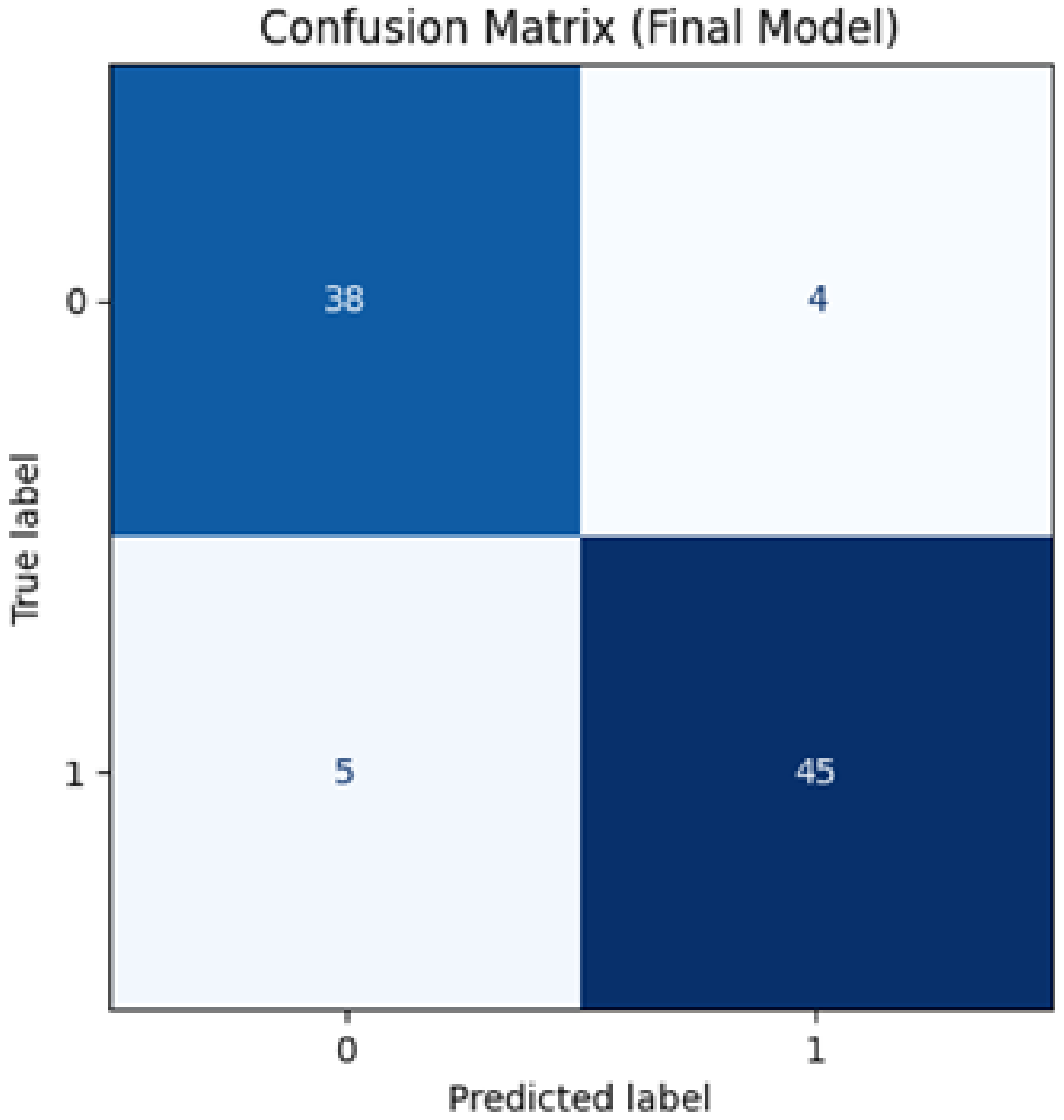

Figure 7

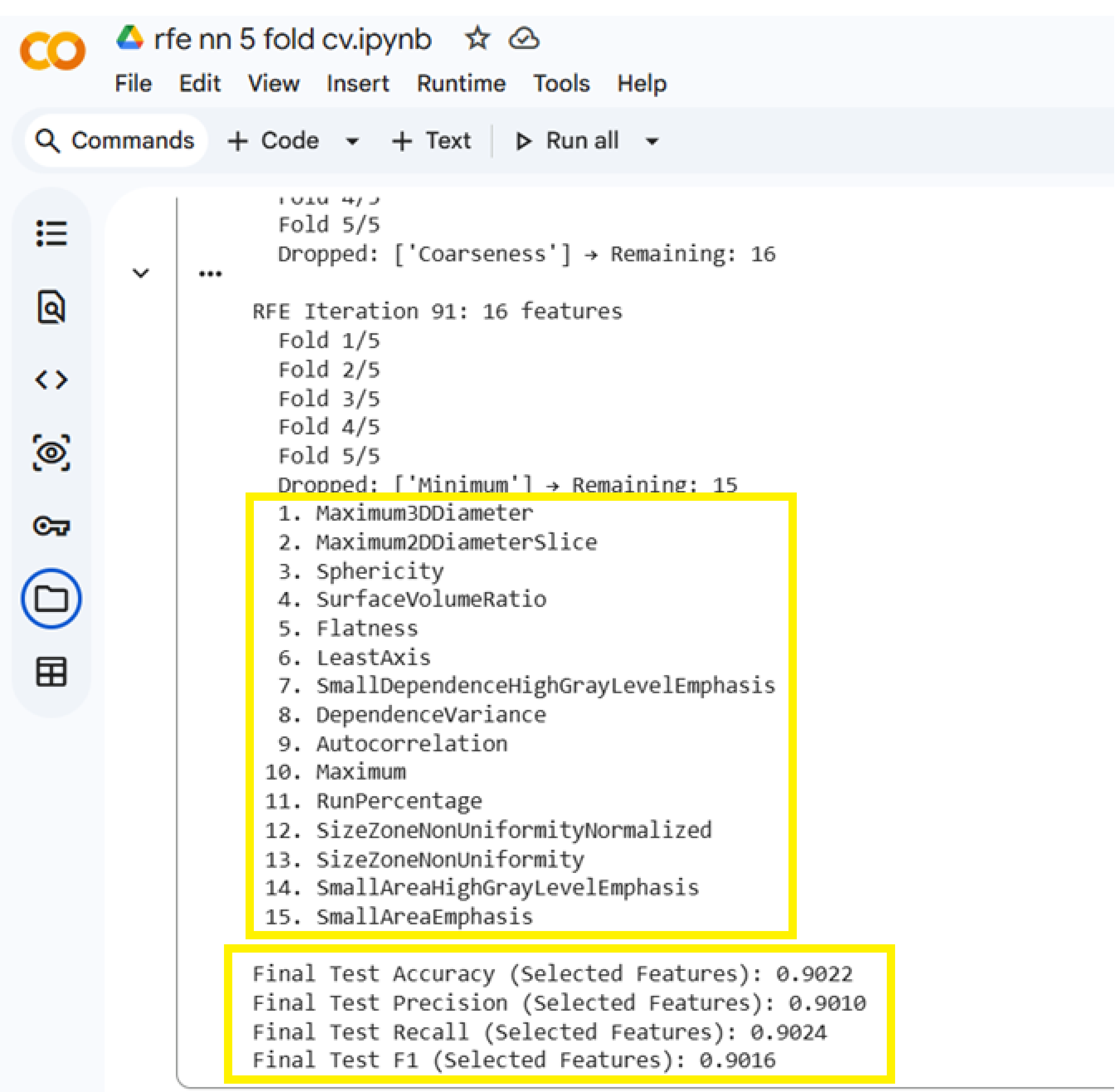
rfe nn 5 fold cv.ipynb
File Edit View Insert Runtime Tools Help
Commands
+ Code
+ Text
Run all
Fold 5/5
Dropped: ['Coarseness'] → Remaining: 16
RFE Iteration 91: 16 features
Fold 1/5
Fold 2/5
Fold 3/5
Fold 4/5
Fold 5/5
Dropped: ['Minimum'] → Remaining: 15
1. Maximum3DDiameter
2. Maximum2DDiameterSlice
3. Sphericity
4. SurfaceVolumeRatio
5. Flatness
6. LeastAxis
7. SmallDependenceHighGrayLevelEmphasis
8. DependenceVariance
9. Autocorrelation
10. Maximum
11. RunPercentage
12. SizeZoneNonUniformityNormalized
13. SizeZoneNonUniformity
14. SmallAreaHighGrayLevelEmphasis
15. SmallAreaEmphasis
Final Test Accuracy (Selected Features): 0.9022
Final Test Precision (Selected Features): 0.9010
Final Test Recall (Selected Features): 0.9024
Final Test F1 (Selected Features): 0.9016

Figure 8

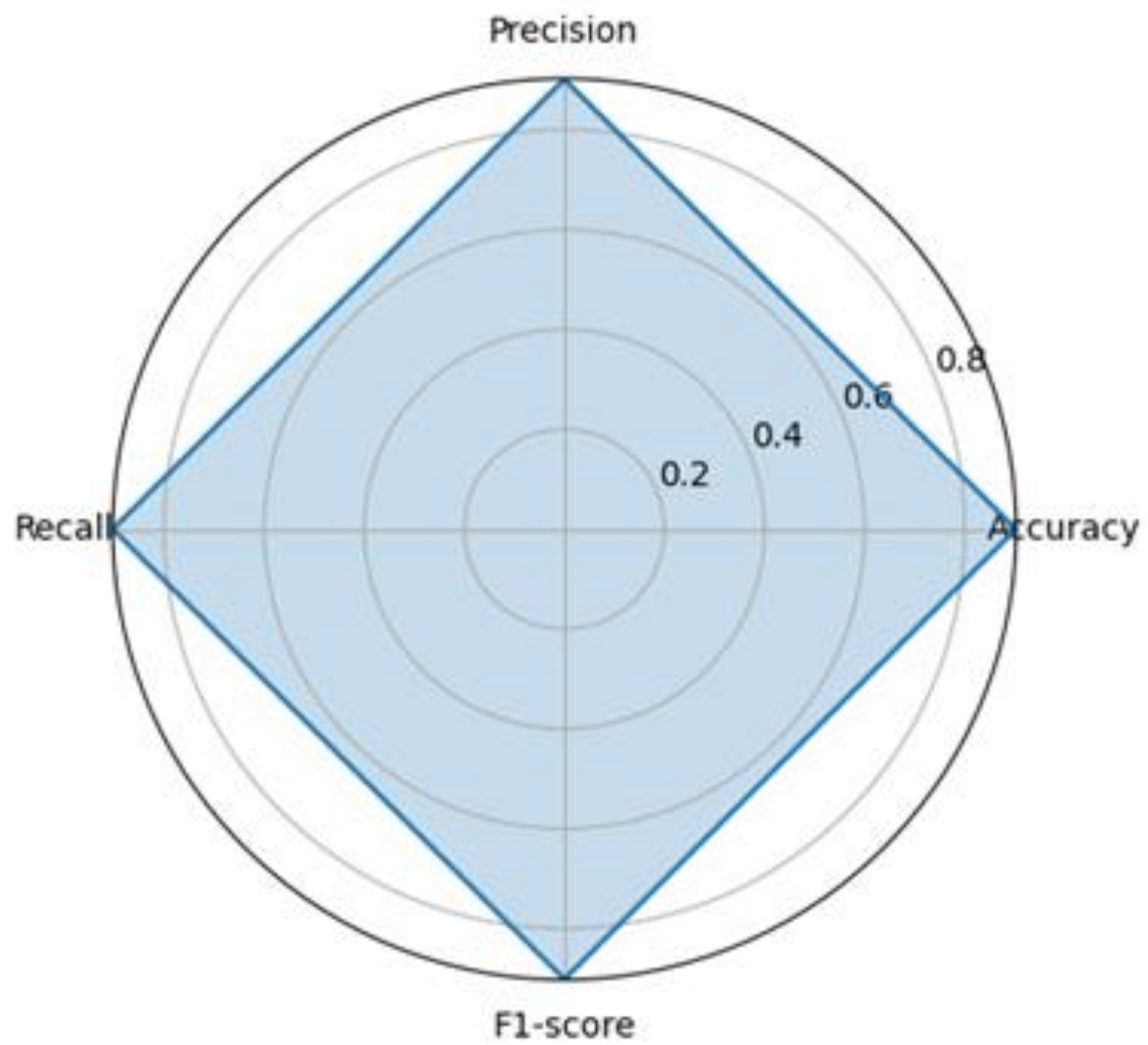

Figure 9

# Table 1

**CT Data Sets Summary (NSCLC Radiomics Dataset)**

| | Before Over Sampling | | After Over-Sampling | |
|---|---|---|---|---|
| | Test Samples | Training Samples | Test Samples | Training Samples |
| Stage I&II | 92 | 42 | 180 | 42 |
| Stage IIIa &IIIb | 238 | 50 | 230 | 50 |

# Table 2

## Hyper-parameters Setting used in Training of MLP and DNN

| **Parameters** | **Values in MLP** | **Values in DNN** |
|---|---|---|
| No. of Classes | 02 (Early Stage and Advanced Stage Cancer) | 02 (Early Stage and Advanced Stage Cancer) |
| Hidden Layers (No. of Neurons ) | 02(128, 64) | 04(128,64,64,64) |
| Activation Layers | ReLU (Hidden layers) Softmax (Output layer) | ReLU (Hidden layers) Softmax (Output layer) |
| Learning Rate | 0.001 | 0.001 |

| Loss Function | Cross-Entropy | Cross-Entropy |
|---|---|---|
| Loss Function | Stochastic Gradient Descent with momentum | Stochastic Gradient Descent with momentum |
| Epochs | 50 | 50 |
| | | |

## Table 3

## COMPUTATIONAL COMPLEXITY COMPARISON OF FEATURE SELECTION METHODS

| Method | Category | Complexity (Approx.) | Key Characteristics | Advantages | Limitations |
|---|---|---|---|---|---|
| ANOVA / Statistical Filtering | Filter | $O(NxF)$ | Evaluates statistical relationship between each feature and class label independently | Very fast and simple | Cannot capture nonlinear feature interactions |
| Mutual Information | Filter | $O(NxF)$ | Measures dependency between features and target variable | Detects nonlinear relations better than ANOVA | Ignores interactions between multiple features |
| LASSO (L1 Regularization) | Embedded | $O(IxNxF)$ | Penalizes feature weights during model training | Produces sparse feature set automatically | Assumes linear relationships |
| SVM-RFE | Wrapper | $O(IxF^2)$ | Iteratively removes features based on SVM weight importance | Good predictive performance | Computationally expensive for large feature sets |
| Random Forest Feature Importance | Embedded | $O(TxNxF)$ | Uses ensemble tree-based importance measures | Handles nonlinear relationships | Importance scores may be biased for correlated features |
| Proposed GL-RFE (Gradient Loss RFE) | Wrapper / Deep Learning | $O(ExNxF)$ | Uses gradients of neural network loss to determine feature importance | Captures nonlinear dependencies and dynamic feature relevance | Requires neural network training and GPU resources |

Here

**N** = number of samples
**F** = number of features
**I** = number of iterations
**T** = number of trees (Random Forest)
**E** = number of training epochs

**Table 4**
**Performance Comparison of GL-RFE with Classical and Deep Learning Techniques**

| Method | Model Type | Accuracy (%) | Precision (%) | Recall (%) | F1-Score (%) |
|---|---|---|---|---|---|
| SVM | Classical ML | 86.20 | 85.90 | 86.50 | 86.10 |
| KNN | Classical ML | 84.75 | 84.10 | 85.00 | 84.50 |
| Logistic Regression | Classical ML | 85.60 | 85.20 | 85.90 | 85.50 |
| Random Forest (RF) | Ensemble ML | 88.00 | 87.00 | 89.00 | 88.00 |
| ANN (MLP) | Deep Learning | 87.00 | 86.00 | 88.00 | 87.00 |
| CNN (Radiomics Input) | Deep Learning | 88.50 | 88.00 | 89.10 | 88.50 |
| Autoencoder + Classifier | Deep Learning | 89.10 | 88.70 | 89.50 | 89.00 |
| **GL-RFE + DNN (Proposed)** | **Deep Learning** | **90.22** | **90.10** | **90.24** | **90.19** |

**Table 5**

**Comparison of GL-RFE with classical Feature Selection techniques**

| Method | Category | Selected Features | Accuracy (%) | Precision (%) | Recall (%) | F1-score (%) |
|---|---|---|---|---|---|---|
| ANOVA (Statistical Filtering) | Filter | 15 | 84.90 | 84.30 | 85.60 | 84.94 |
| Mutual Information | Filter | 15 | 85.80 | 85.10 | 86.40 | 85.74 |
| Chi-Square | Filter | 15 | 84.50 | 83.90 | 85.20 | 84.54 |
| LASSO (L1 Regularization) | Embedded | 15 | 86.70 | 86.10 | 87.30 | 86.69 |
| Random Forest Importance | Embedded | 15 | 87.90 | 87.40 | 88.50 | 87.94 |
| Sequential Feature Selection (SFS) | Wrapper | 15 | 86.75 | 86.10 | 87.20 | 86.64 |
| Recursive Feature Elimination (RFE) | Wrapper | 15 | 88.30 | 87.95 | 88.80 | 88.37 |
| ANN (All Features – 106) | Deep Learning | 106 | 87.10 | 86.85 | 87.90 | 87.37 |
| **Proposed GL-RFE + DNN** | Wrapper / DL | **15** | **90.22** | **90.10** | **90.24** | **90.19** |